\documentclass[a4paper, 10 pt, conference]{hsmr} 
\usepackage[utf8]{inputenc}
\IEEEoverridecommandlockouts                       
\usepackage{mathtools}
\usepackage{geometry}
\usepackage[labelsep=space]{caption}
\usepackage{newtxmath,newtxtext}
\usepackage{authblk}
\usepackage{pgfplots}
\usepackage{todonotes}
\usepackage{subfig}
\usepackage{graphicx}

\usepackage{newtxtext,newtxmath}

\pgfplotsset{compat=newest} 
 
\title{\LARGE \bf
Collaborative Robotic Ultrasound Tissue Scanning for Surgical Resection Guidance in Neurosurgery}

\author{\LARGE Alistair Weld*$^{1}$}
\author{\LARGE Michael Dyck*$^{2,3}$} 
\author{\LARGE Julian Klodmann$^{2}$} 
\author{\LARGE Giulio Anichini$^{4}$}
\author{\LARGE \\Luke Dixon$^{4}$} 
\author{\LARGE Sophie Camp$^{4}$}
\author{\LARGE Alin Albu-Schäffer$^{2,3}$}
\author{\LARGE Stamatia Giannarou$^{1}$}

\affil{\Large\textit{$^{1}$ Hamlyn Centre for Robotic Surgery, Imperial College London,}\\ \Large\textit{$^{2}$Institute of Robotics and Mechatronics, German Aerospace Center,}\\ \Large\textit{$^{3}$Department of Informatics, Technical University of Munich,}\\
\Large\textit{$^{4}$Department of Neurosurgery, Charing Cross Hospital, Imperial College London, UK}\\
\Large\textit{a.weld20@imperial.ac.uk, michael.dyck@dlr.de}}

\begin{document}

\maketitle
\thispagestyle{empty}
\pagestyle{empty}

\section*{INTRODUCTION}
\let\thefootnote\relax\footnote{*These authors contributed equally to the work. This project was supported by UK Research and Innovation (UKRI) Centre for Doctoral Training in AI for Healthcare (EP/S023283/1), the Royal Society (URF$\setminus$R$\setminus$2 01014]), the NIHR Imperial Biomedical Research Centre and the International Graduate School of Science and Engineering (IGSSE); TUM Graduate School}
The main goal of surgical oncology is to achieve complete resection of cancerous tissue with minimal iatrogenic injury to the adjacent healthy structures. Brain tumour surgery is particularly demanding due to the eloquence of the tissue involved. There is evidence that increasing the extent of tumour resection substantially improves overall and progression-free survival. Realtime intraoperative tools which inform of residual disease are invaluable.
Intraoperative Ultrasound (iUS) has been established as an efficient tool for tissue characterisation during brain tumour resection in neurosurgery \cite{iUS}. 

The integration of iUS into the operating theatre is characterised by significant challenges related to the interpretation and quality of the US data. 
The capturing of high-quality US images requires substantial experience and visuo-tactile skills during manual operation. Recently, robotically-controlled US scanning systems have been proposed (see e.g. \cite{rus}) but the scanning of brain tissue poses major challenges to robotic systems because of the safety-critical nature of the procedure, the very low and precise contact forces required, the narrow access space and the large variety of tissue properties (hard scull, soft brain structure).

The aim of this paper is to introduce a robotic platform for autonomous iUS tissue scanning to optimise intraoperative diagnosis and improve surgical resection during robot-assisted operations. To guide anatomy specific robotic scanning and generate a representation of the robot task space, fast and accurate techniques for the recovery of 3D morphological structures of the surgical cavity are developed. The prototypic DLR MIRO surgical robotic arm \cite{miro} is used to control the applied force and the in-plane motion of the US transducer. A key application of the proposed platform is the scanning of brain tissue to guide tumour resection.

\section*{MATERIALS AND METHODS}
\textbf{Intraoperative Surgical Navigation.} The 3D reconstruction of surgical scenes is uniquely challenging due to textureless surfaces, occlusions, specular highlights and tissue deformation. To deal with these challenges, we develop a novel deep learning approach that utilises stereo vision for geometric focused depth estimation. Our method advances conventional 3D reconstruction approaches which are based on the direct regression of depth information, by developing a neural network which learns to replicate the behaviour of a structured light projector. Geometric awareness is enforced through the learning of the reflection of light off surfaces.

A Unet architecture has been designed to take a stereo image as input and to predict how the projected structured light should appear on the surfaces on the left and right images, separately. Specifically, we are predicting binary code structured light, which consists of vertical bars of white and black colour. To predict these binary patterns, we purpose our neural network to be a Sigmoid binary classifier. So the neural network should be predicting whether or not a pixel should have a value of 1 or 0, where white corresponds to the value 1 and black to the 0. These predicted binary codes can then be processed into disparity or depth maps, using 2D cross correlation over the epipolar lines of the stereo images.

\textbf{Collaborative Robotic Tissue Scanning.} To fulfil the safety critical requirements of robotic iUS scanning and to cope with the low stiffness of brain tissue we develop an impedance controller using task-specific coordinates defined with respect to the surface of the anatomy of interest. The recovered 3D morphological structure is utilised to represent the anatomical surface as a triangular mesh. We define the task coordinates as the orientation of the US probe axis to coincide with the surface normal and its distance $d$ to the surface. Controlling this distance allows us to realise different scenarios of probe-tissue interaction, such as contact avoidance with a safety margin for re-scanning and updating the 3D structure of the tissue surface, contact establishment, or interaction with a desired contact force. We obtain a planar representation of the anatomical surface, described by 2D $(u, v)$-coordinates. Scanning trajectories along the surface can be planned in 2D and automatically executed. The $(u, v)$-coordinates with zero impedance allows the clinician to move along the surface, while the controller is taking care of probe orientation and contact forces.

Controlling the robot at different distances to the tissue implicitly creates a foliation of the Cartesian space, parallel surfaces, obtained by shifting the original tissue surface along its normal direction. We define the planar $(u, v)$-coordinates in a novel way that allows for consistent impedance control along and in-between all leaves of the foliation. This novel concept allows to control and continuously adapt desired interaction forces between US probe and tissue, capable of safe US scanning on tissues as soft as the human brain. Due to the use of the unified impedance control framework in \cite{imp}, passivity and stability are guaranteed.

\section*{RESULTS}
\begin{figure}
\centering
\includegraphics[width=0.8\columnwidth]{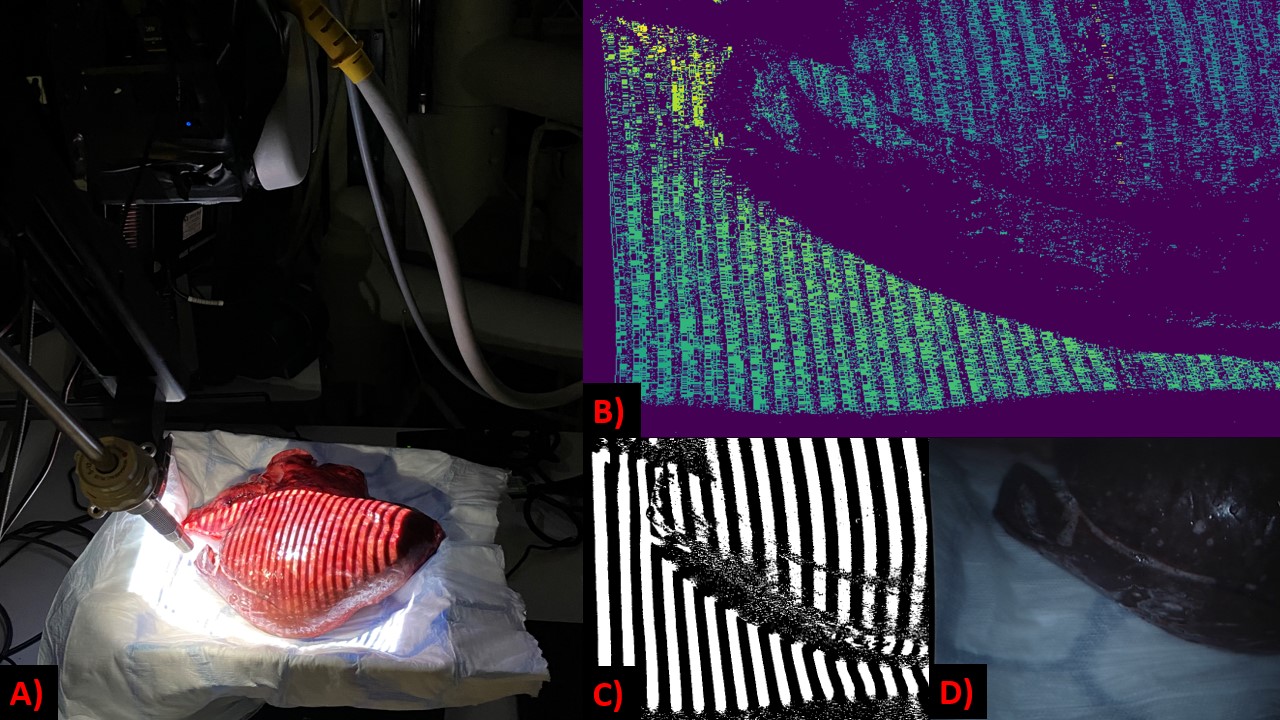}
\caption{A) camera, projector setup with sheep's kidney B) generated depth map C) binarised structured light pattern D) laproscope image.}
\label{fig_example}
\end{figure}

Due to the lack of publicly available datasets containing binary code structured light patterns, our proposed model has been trained and tested on simulated data of real objects, created with VisionBlender \cite{visionblender}. Results so far have shown successful depth estimation, without requiring training on depth ground truth. The validation mean absolute error was $1.4\ pixels$, where the mean ground truth disparity of the objects is $16.1\ pixels$. Early comparison to direct depth regressing networks, has indicated to an improved robustness to pixel perturbation (brightness and contrast) by up to $50~\%$, using our technique. Our model will be further validated on a dataset which we are currently developing with images of ex vivo and medical phantoms as shown in Fig. \ref{fig_example}. 

\begin{figure}
\centering
\includegraphics[width=\columnwidth]{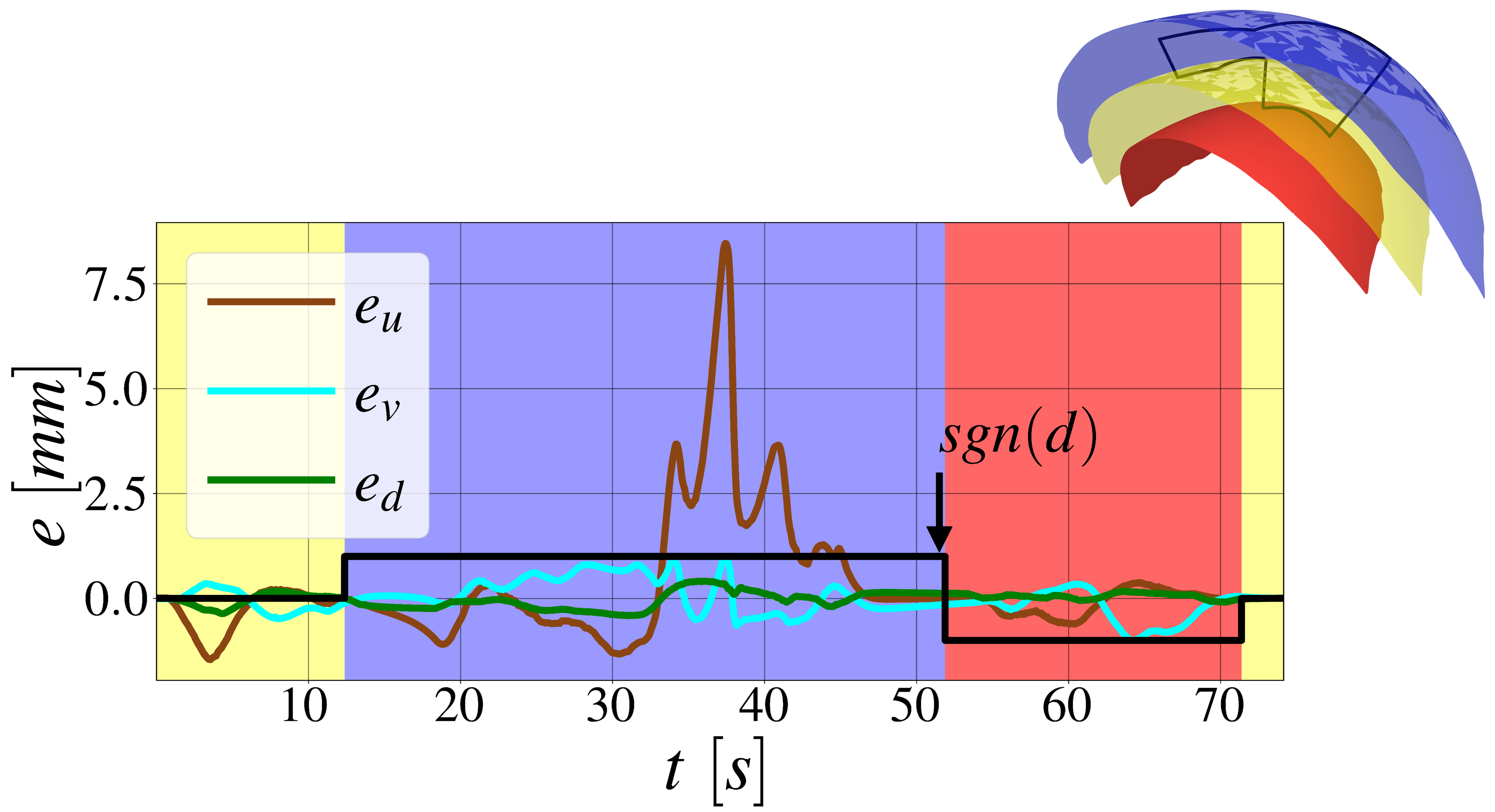}
\caption{Error $e$ in $(u, v)$-coordinates, distance $d$ for trajectory (upper right) along tissue surface (yellow, $d_{des}=0~mm$), positive leaf (blue, $d_{des}=50~mm$) and negative leaf (red, $d_{des}=-50~mm$). The black graph depicts the signum function of the desired distance $d$, where $sgn(d)=\pm1$ corresponds to $d_{des}=\pm50~mm$, respectively.}
\label{results_control}
\end{figure}

Fig. \ref{results_control} shows the performance of the developed impedance controller in a realistic simulation, tracking a trajectory along a discretised surface patch ($d=0$, establish contact), a positive leaf ($d>0$, avoid contact) and negative leaf ($d<0$, apply a particular force), with the corresponding control errors ($e_u,\ e_v,\ e_d$). We demonstrate consistency of the planar $(u, v)$-coordinates and accurate tracking behaviour along all leafs, with a root mean squared error (RMSE) in the controlled distance of $RMSE_d = 0.369~mm$. The higher errors in $u$ most likely arise from dynamic coupling and lack of feedforward control terms in the presented experiment. 

\section*{DISCUSSION}
In this paper, we have presented a robotic platform for iUS tissue scanning which will optimise diagnosis with the aim of improving both the efficacy and safety of tumour resection. The development of the above technologies will have a significant impact on the surgeon’s sensing, completeness and safety during tumour resection and ultimately on the management of the patient. Our future work will focus on further improving the control, as well as validating the proposed visual servoing setup on ex vivo and medical phantoms. 

\nocite{*}
\bibliographystyle{IEEEtran}
\bibliography{HSMR}

\end{document}